\documentclass[letterpaper, 10 pt, conference]{ieeeconf}  

\IEEEoverridecommandlockouts                              

\usepackage{graphics} 
\usepackage{epsfig} 
\usepackage{times} 
\usepackage[pagebackref,breaklinks,colorlinks]{hyperref}
\usepackage{subfigure}
\usepackage{xcolor}
\usepackage{color}
\usepackage{colortbl}
\usepackage{multicol}
\usepackage{multirow}
\usepackage{makecell}
\usepackage{amsmath}
\usepackage{epstopdf}
\usepackage{bbding}
\usepackage{bbm}
\usepackage{bm}
\usepackage{url}            
\usepackage{booktabs}       
\usepackage{amsfonts}       
\usepackage{nicefrac}       
\usepackage{microtype}      

\usepackage{amsmath,amssymb}
\let\mathcal\undefined
\DeclareMathAlphabet{\mathcal}{OMS}{cmsy}{m}{n}

\usepackage[ruled,linesnumbered]{algorithm2e}
\SetKwComment{Comment}{/* }{ */}

\title{\LARGE \bf
Online Monocular Lane Mapping Using Catmull-Rom Spline
}

\author{Zhijian Qiao, Zehuan Yu, Huan Yin and Shaojie Shen            
\thanks{This work was supported in part by the HKUST Postgraduate Studentship, in part by the HKUST-DJI Joint Innovation Laboratory, and in part by the Hong Kong Center for Construction Robotics (InnoHK center supported by Hong Kong ITC).}
\thanks{The authors are with the Department of Electronic and Computer Engineering, The Hong Kong University of Science and Technology, Hong Kong, China. E-mail: zqiaoac@connect.ust.hk, zyuay@connect.ust.hk, eehyin@ust.hk, eeshaojie@ust.hk}
\thanks{Corresponding author: Huan Yin}
}

\begin{document}
\maketitle
\thispagestyle{empty}
\pagestyle{empty}

\begin{abstract}
In this study, we introduce an online monocular lane mapping approach that solely relies on a single camera and odometry for generating spline-based maps. Our proposed technique models the lane association process as an assignment issue utilizing a bipartite graph, and assigns weights to the edges by incorporating Chamfer distance, pose uncertainty, and lateral sequence consistency. Furthermore, we meticulously design control point initialization, spline parameterization, and optimization to progressively create, expand, and refine splines. In contrast to prior research that assessed performance using self-constructed datasets, our experiments are conducted on the openly accessible OpenLane dataset. The experimental outcomes reveal that our suggested approach enhances lane association and odometry precision, as well as overall lane map quality. We have open-sourced our code\footnote{\href{https://github.com/HKUST-Aerial-Robotics/MonoLaneMapping}{https://github.com/HKUST-Aerial-Robotics/MonoLaneMapping}} for this project.
\end{abstract}

\section{INTRODUCTION}

High-definition (HD) map is critical for autonomous driving. As an essential element in the HD map, lane markings play a vital role in vehicle pose estimation, trajectory planning, and decision-making for high-level tasks. However, creating and maintaining lane-contained HD maps typically involve multiple processes~\cite{homayounfar2019dagmapper,tang2022thma}, e.g., high-quality data collection by specialized vehicles, 3D reconstruction and landmark annotations. Such expensive mapping systems hinder the promotion of autonomous driving in large-scale environments. 

One promising direction is to achieve online lane mapping using cheap sensors on standard vehicles. One might suggest obtaining lane maps by directly accumulating lane detection results using vehicle odometry information without designing extra modules. This approach will face two significant problems. First, the lane maps obtained from cumulative semantic point clouds do not have instance-level information. Second, odometry drift and detection errors can make lane markings blurred and inaccurate. Thus, offline manual annotation and vectorization are still required if using accumulated lane detection results.

In this study, we propose to use simultaneous localization and mapping (SLAM) techniques to achieve online lane mapping. However, online lane mapping remains a challenging task because lane markings are often curved and more difficult to represent and optimize than the conventional point, line, and plane representations in the SLAM community. To tackle these challenges, we propose to use Catmull-Rom spline \cite{catmull1974class} to model the lane markings (also called lane for brevity in the following text). We choose the Catmull-Rom spline mainly for three reasons. First, it naturally has $C^1$ continuity and is smooth enough for lane marking modeling. Second, it is a lightweight representation compared to point-set representation; thus, we only need to store control points to build a continuous lane-marking model. Thirdly, all of the control points are situated on the spline curves, which facilitates the acquisition of an optimal control point initialization and enables the development of a coarse-to-fine spline parameterization technique.

\begin{figure}[tbp]
	\centering
	\includegraphics[width=0.9\linewidth]{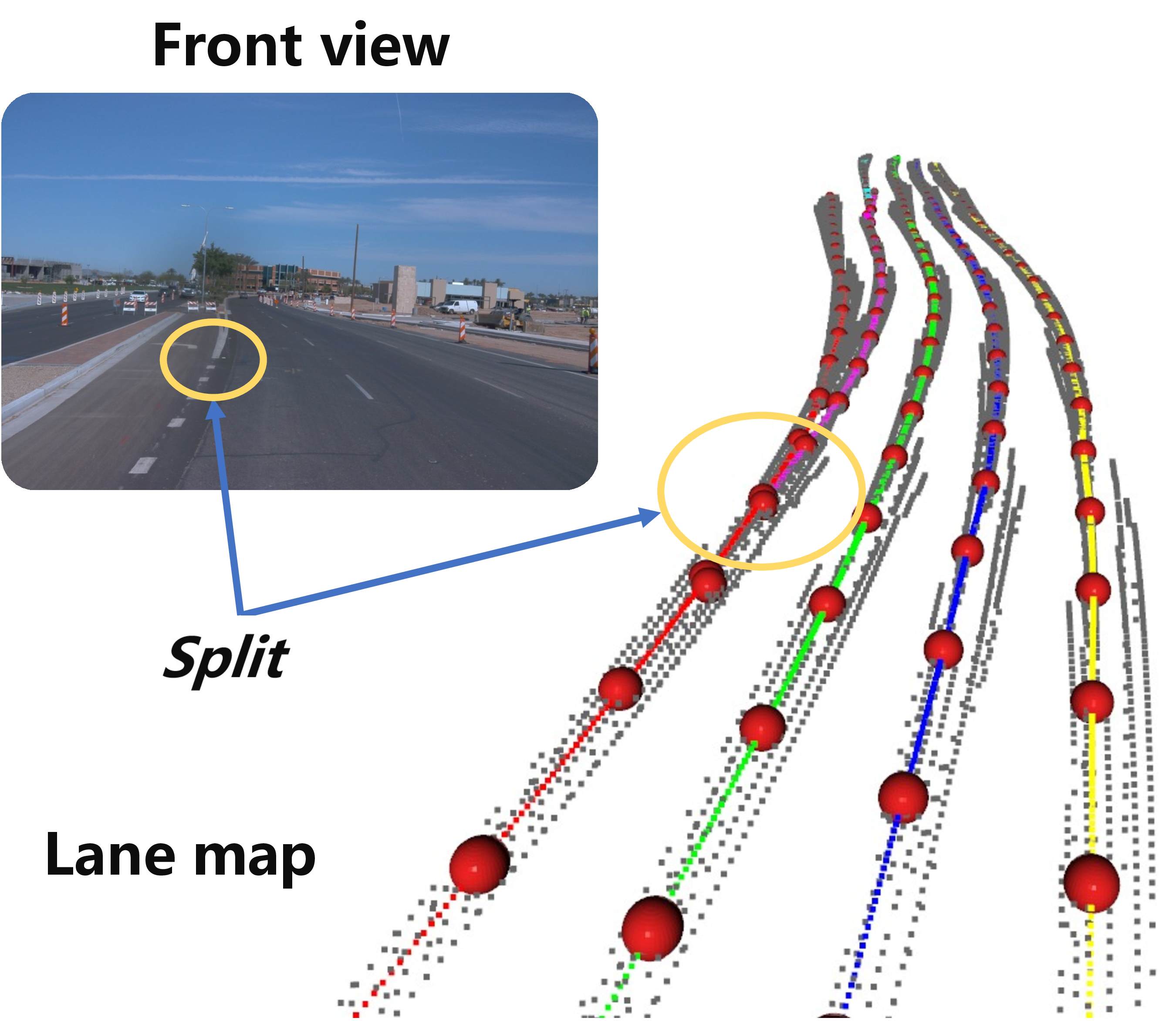}
	\caption{Experimental results on the OpenLane dataset. The gray points represent the results accumulated with odometry for multi-frame detection. The colored curves represent the sampling points of the splines for different instances in the map. The red spheres represent the control points of the splines.}
	\label{figure:cover}
	\vspace{-0.8cm}
\end{figure}

A complete online lane mapping system is designed based on the Catmull-Rom spline representation, as shown in Fig. \ref{figure:cover}. The proposed system allows autonomous vehicles to construct local lane maps in real-time using temporal images and odometry data, which can be utilized for self-localization, planning, and crowd-sourcing update~\cite{pannen2020keep}. Overall, the specific contributions can be summarized as follows:
\begin{itemize}
    \item We present an online monocular lane mapping system that includes lane tracking and map optimization subsystems. The system can directly output lightweight instance-level lane maps represented by Catmull-Rom splines without offline vectorization.
    \item Each part of the system is meticulously designed to combine the properties of lane markings and splines, including lane association, pose estimation, spline initialization, extension, and optimization.
    \item Experiments on the publicly available dataset OpenLane show that our proposed method can improve lane association, odometry accuracy, and map quality.
\end{itemize}

\begin{figure*}[t]
	\centering
	\includegraphics[width=\linewidth]{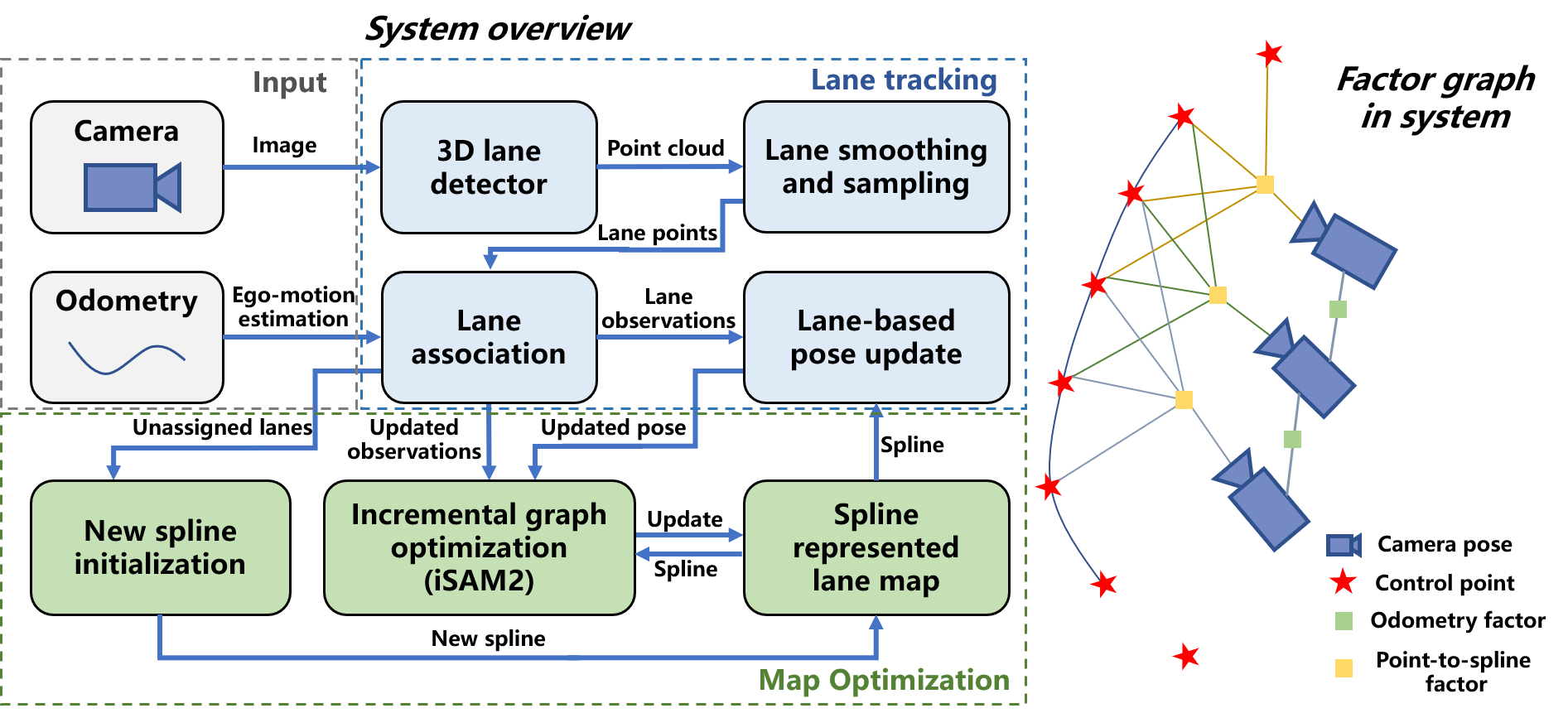}
	\caption{The block diagram illustrates the whole pipeline of the proposed monocular lane mapping system. The system is divided into two parts: lane tracking and map optimization. The former is for lane marking association and pose update, while the latter is for spline initialization, extension and optimization. The factor graph is shown on the right. Unlike the traditional binary visual factor, a point-to-spline factor is involved in optimizing the four control point landmarks. }
	\label{figure:system}
	\vspace{-0.4cm}
\end{figure*}
\section{RELATED WORK}

\subsection{3D Lane Detection}
In recent years, substantial advancements in accuracy and speed have been made in the field of online visual lane detection~\cite{qin2022ultra, bar2014recent}. The majority of these methodologies emphasize the detection of 2D lanes in image space. Contrary to points, 2D lanes present challenges when projected into the world coordinate system through triangulation. Rather, most studies utilize inverse perspective mapping (IPM)~\cite{jeong2016adaptive} to cast lanes onto the road surface. However, this approach is heavily dependent on the assumption of flat ground and the estimation of the camera pitch angle, factors which can lead to distortions in 3D lanes during vehicle vibrations or changes in road height~\cite{zhou2022visual}.

A distinct methodology entails manipulating the features derived from images in the Bird's-Eye-View (BEV) space and detecting the lane markings, as illustrated by HDMapNet~\cite{li2022hdmapnet} and MapTR~\cite{liao2022maptr}, or directly recognizing the lane graph from the BEV image~\cite{zurn2021lane}. However, such methods do not regress the heights. In contrast, Persformer~\cite{chen2022persformer}, the open-sourced cutting-edge method for 3D lane detection, regresses offsets and heights by establishing 3D lane anchors on BEV space, thereby directly predicting 3D lanes.

\subsection{Lane Mapping}
Lane mapping has been extensively studied in the last decade because of its importance for self-driving cars. Works in ~\cite{cheng2022vision,jeong2017road} have used semantic segmentation combined with depth or flat ground to reconstruct a 3D lane point cloud. However, accumulating the point clouds with poses makes the lane map size large. Researchers in~\cite{paz2020probabilistic,qin2021light} proposed to create rasters to reduce the map size, but the rasters also introduce quantization errors. A better approach is to approximate the lane using polylines and estimate each node's position and uncertainty~\cite{heidenreich2015laneslam,zhou2022visual}. However, this representation is clearly insufficient in some curved lane scenarios. Another option is the spline~\cite{ali2021road,meier2018visual}, which is compact and naturally continuous, with the disadvantage that it is challenging to perform control point initialization, extension, and optimization. Even though spline-aided LiDAR mapping methods~\cite{usman2019extensive,pedraza2009extending}  have been proposed, their methods cannot be directly used for monocular lane mapping because they rely on high-precision lidar measurements.

In addition, several deep learning-based online mapping methods~\cite{li2022hdmapnet,liao2022maptr} show impressive results. Still, few works implement temporal fusion to improve the consistency and accuracy of the maps.

We do not need LiDAR or planar ground assumptions compared with previous works. Instead, we directly predict 3D lane markings from a monocular camera. To achieve this, our method utilizes Catmull-Rom splines, enabling efficient modeling and optimization of the complex and often curved nature of lane markings.

\subsection{Edge-based VO}
\label{sec:edge_vo}
Our approach is also partly inspired by edge-based VO. Edges were first introduced into the VO pipeline to avoid falling into local optima and be more robust to illumination~\cite{kneip2015sdicp,ling2018edge}. Nearest neighbor search for edges can ignore the assumption of basic brightness constancy. Early edge-based VO represents image edges as a set of pixels for which a distance field is created to obtain the residuals from the map projected points to the edges and optimize the pose~\cite{kneip2015sdicp,tarrio2015realtime}. However, there are three problems with this. First, distance fields can cause non-differentiability in the optimization process. For this reason, Ling \textit{et al.}~\cite{ling2018edge} proposed to use the sub-gradient method. Zhou \textit{et al.}~\cite{zhou2018canny} proposed using the nearest neighbor field instead of the distance field. Second, researchers in ~\cite{nurutdinova2015towards} pointed out two ways of establishing residuals, data-to-model and model-to-data. At the same time, the latter may cause partial observations in the case of partial occlusion. In this case, Zhou \textit{et al.}~\cite{zhou2018canny} still used the former because it claims that the proposed point-to-tangent residuals can reduce this effect. Also, if the model refers to a spline and data for points. Partial observations also originate from the difference between discrete and continuous representations. Third, there are no natural 3D edges in the 3D world, meaning that the edges detected on the image are wrong, such as apparent contours or occlusions~\cite{wang2018fully}.

In this study, considering the characteristic of Catmull-Rom splines, we design a coarse-to-fine parameterization method to find the point-to-curve nearest neighbor (aka foot point), compute the residuals, and use the data-to-model way. Moreover, the detected lane is natural and unique as a semantic edge.

\section{METHOD}
To better describe our proposed system, we first present a system overview and then provide comprehensive illustrations on subsystems in the following subsections.

\subsection{System Overview}

The structure of the proposed monocular lane mapping system is presented in Fig.~\ref{figure:system}. The system outputs compact lane marking maps represented as splines using only monocular cameras and odometry (e.g., VIO, LIO) as input, without needing a priori navigation maps or aerial photographs. Specifically, the proposed framework consists of two subsystems: lane tracking and map optimization. A neural network predicts 3D lane markings directly based on the input image in lane tracking. Then the predictions are further processed to match the subsequent needs, presented in Sec.~\ref{subsec:ldr}. Subsequently, the processed lane markings are associated with the lane markings in the map in combination with the poses provided by the odometry (Sec.~\ref{subsec:la}). Finally, the poses are updated based on the association results (see Sec.~\ref{subsec:pu}). In map optimization, the splines are first initialized from scratch, or the original splines are extended based on the newly obtained detection results (Sec.~\ref{subsec:li}). Finally, an incremental optimization framework, iSAM2~\cite{kaess2012isam2}, is applied to add new observations to incrementally update the splines in the map without losing information from past observations (Sec.~\ref{subsec:so}).

\subsection{Lane tracking}
\subsubsection{Lane Representation}
\label{subsec:ldr}
\begin{figure}[t]
	\centering
	\subfigure[]{	
        \label{figure:coeffs}
		\includegraphics[width=4.1cm]{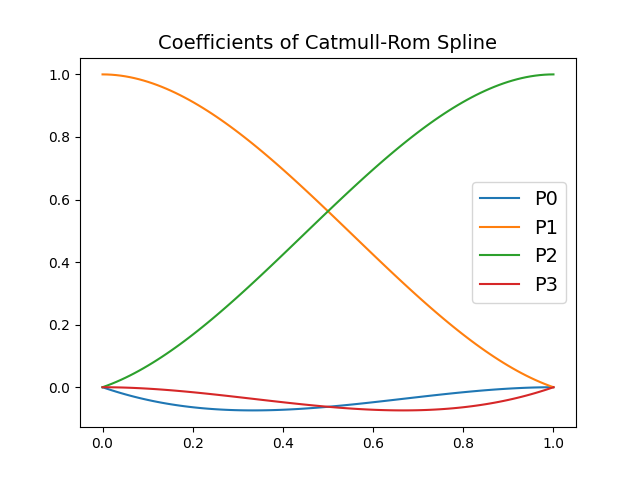}} 
	\subfigure[]{	
	    \label{figure:parameterization}
		\includegraphics[width=4.1cm]{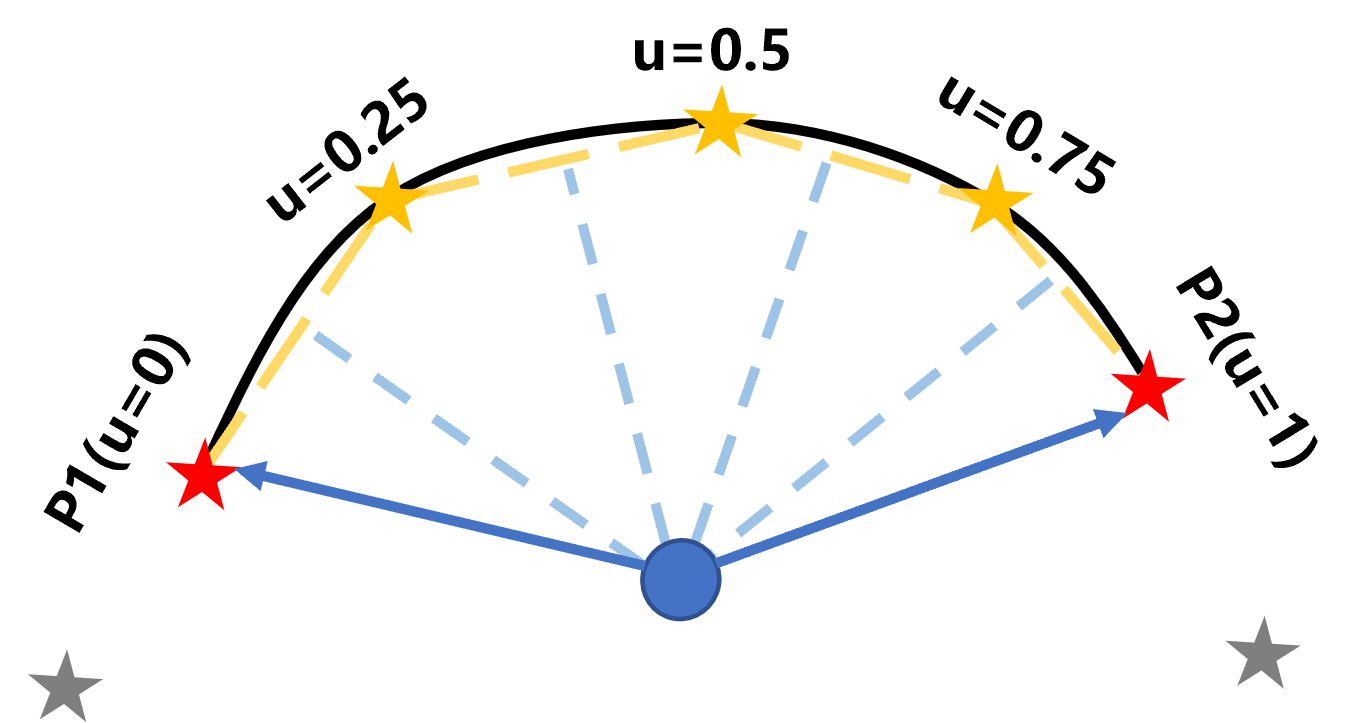}}
	\caption{(a) Coefficients of four control points for $\tau=\frac{1}{2}$. (b) Coarse-to-fine parameterization. Red stars represent control points, and yellow for sampled waypoints. We first find the two closest control points and then determine the parameters by finding the foot point on the polyline. Note that the curvature of the actual lane markings is small and not as large as in the figure.}
	\vspace{-0.2cm}
\end{figure}

In this research, Persformer~\cite{chen2022persformer} is utilized to attain lane detection results, which include unordered lane marking points and their associated instance-level labels. Moreover, we distinguish the representation of lanes concerning observations (detection) and landmarks (maps). Particularly, for network predictions, taking into account their sparsity and noise, they are initially converted to a local reference frame (LRF) where the primary direction of the lane is aligned with the X-axis. Subsequently, a cubic polynomial is fitted to both X-Y and X-Z coordinates and sampled at a specific resolution (established as 0.5m in our experiments). As a result, a lane observation is written as,
\begin{equation}
    D_i=\left\{{ }^d p_{1: M}, \mathrm{f}_{xy}, \mathrm{f}_{xz}, c, { }^d \sigma_{1: M} \right\}
\end{equation}
where ${ }^d p_{1: M} \in \mathbb{R}^3 $ is the sampling point on the detected lane, $\mathrm{f}_{xy}$ and $\mathrm{f}_{xz}$ are the polynomial coefficients, respectively, and $c$ is the category (e.g., double yellow line, white solid line). ${ }^d \sigma_{1: M}$ is the detection noise's standard deviation, which can be set to be proportional to the 2-norm of the point.

The lane landmark in the map is then represented by the Catmull-Rom spline, written as $L_j=\left\{P_{0: N+1}, c\right\}$, where $P_{0: N+1}$ are the control points in the spline, and $c$ is the category. The piecewise spline curve $L_j$ has N segments (every two segments share three control points), and each segment $l$ has four adjacent control points that determine the curves' points ${ }^l p(u)$:
\begin{equation}
{ }^l p(u)=\left[\begin{array}{c}
1 \\
u \\
u^2 \\
u^3
\end{array}\right]^{\mathrm{T}}\left[\begin{array}{cccc}
0 & 1 & 0 & 0 \\
-\tau & 0 & \tau & 0 \\
2 \tau & \tau-3 & 3-2 \tau & -\tau \\
-\tau & 2-\tau & \tau-2 & \tau
\end{array}\right]\left[\begin{array}{c}
{ }^l P_0^{\mathrm{T}} \\
{ }^l P_1^{\mathrm{T}} \\
{ }^l P_2^{\mathrm{T}} \\
{ }^l P_3^{\mathrm{T}}
\end{array}\right]
\end{equation}
where $u \in \left[0,1\right]$ is called the \emph{parameter}. The process of finding $u$ corresponding to a point on the curve is called \emph{parameterization}; $\tau$ controls how sharply the curve blends and often is set to $\frac{1}{2}$. Without loss of generality, we refer to these four control points as $\left[{ }^l P_0, { }^l P_1, { }^l P_2, { }^l P_3\right]$. In the later sections, $l$ and $d$ may be omitted for brevity. In this case, a point on the curve can be viewed as a weighting of four control points. The coefficients are shown in Fig.~\ref{figure:coeffs}.

\subsubsection{Lane association}
\label{subsec:la}

Given a set of detection $\mathbb{D}$ and a set of landmark $\mathbb{L}$, lane association aims to match detection $D_i$ to an existing landmark $L_j$ or to identify a new lane. For this purpose, we formulate the problem as an assignment problem using a bipartite graph and solve it with the K-M algorithm~\cite{kuhn1955hungarian,munkres1957algorithms}, while the critical point is how to determine the edges and their weights. Firstly, the associated lanes should have the same category. Another natural idea is that one can first sample some points on the spline and then calculate the distance between two point clouds, such as the Chamfer distance, to determine the similarity of $D_i$ to $L_j$. However, the Chamfer distance can always be computed, so a vertex in the bipartite graph will have an edge, resulting in no new lane markings being generated. For this reason, inspired by recent work KISS-ICP~\cite{vizzo2023kiss} and the work by Kim \textit{et al.}~\cite{kim2021hd}, we constrain the search range as well as the upper bound of Chamfer distance.

We compute the difference between the odometry and the true pose $\left( \Delta R, \Delta t\right)$. Unlike the conventional assumption of noise in the Lie group tangent space, we use only two parameters, the standard deviation of rotation and translation, $\sigma_\theta$ and $\sigma_t$, respectively, where: 
\begin{equation}
    \theta = \arccos \left(\frac{\operatorname{tr}(\Delta R)-1}{2}\right), t=\|\Delta t\|_2
\end{equation}
Thus for a point, $p_k$ in $D_i$, the upper bound $\delta_k$ (two-sigma rule with 95\%) on its distance from the true matching point can be written as,
\begin{equation}
    \delta_k = 2\| p_k\|_2\sin{\frac{2\sigma_\theta}{2}} + 2\sigma_t + 2{}^d \sigma_k
\end{equation}

Hence, given the sampled points on the spline $q_{1:Q}$, detection points $p_{1:M}$, and the odometry pose $T$, we redefine the distance between $D_i$ and $L_j$ as
\begin{align}
d_{ji}=\sqrt{\frac{M}{n_a}} \frac{1}{n_a} \sum_{k=1}^M \mathbb{I}\left(d_k<\delta_k\right)d_k, 
d_k=\left\|Tp_k-q_{k^\prime}\right\|_2
\end{align}
where $n_a$ represents the number of points satisfying the distance threshold, $\mathbb{I}$ is the indicator function, $q_{k^\prime}$ is the neareast sampled point to $p_k$, and $\sqrt{\frac{M}{n_a}}$ is multiplied to penalize a low match rate. Further, we set an upper bound $\sqrt{2}mean(\delta_k)$ for $d_{ji}$ to determine if a new lane appears. $\sqrt{2}$ means that at least half of the points are matched. 

\begin{figure}[tbp]
	\centering
	\includegraphics[width=\linewidth]{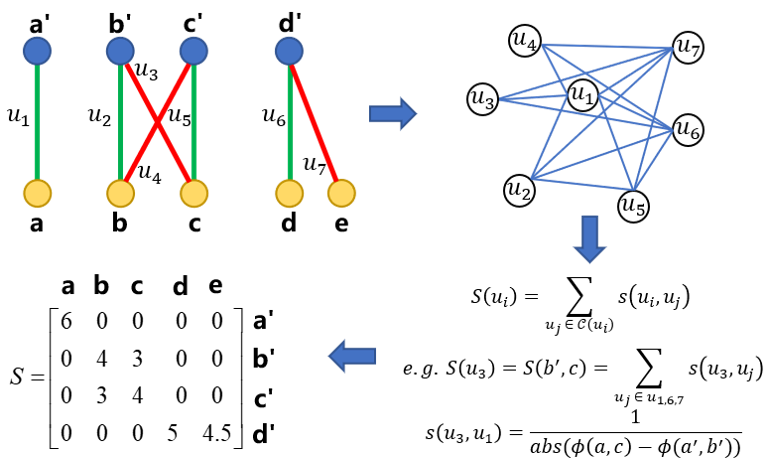}
	\caption{(a) Lane marking detection is represented by yellow points, and lane markings in the map are represented by blue points. The $u$ represents the association obtained from the Euclidean distance, where the red color is wrong. (b) An undirected graph is created, and the vertex represents the association. If two associations have lateral sequence consistency, they have an edge. (c) The definition of the weights of each edge is given. The degree of a vertex is the sum of the edges. (d) The degrees of the vertices will be used as edges of the bipartite graph in the assignment problem.}
	\label{figure:consistency}
	\vspace{-0.4cm}
\end{figure}

\begin{figure}[tbp]
	\centering
	\includegraphics[width=\linewidth]{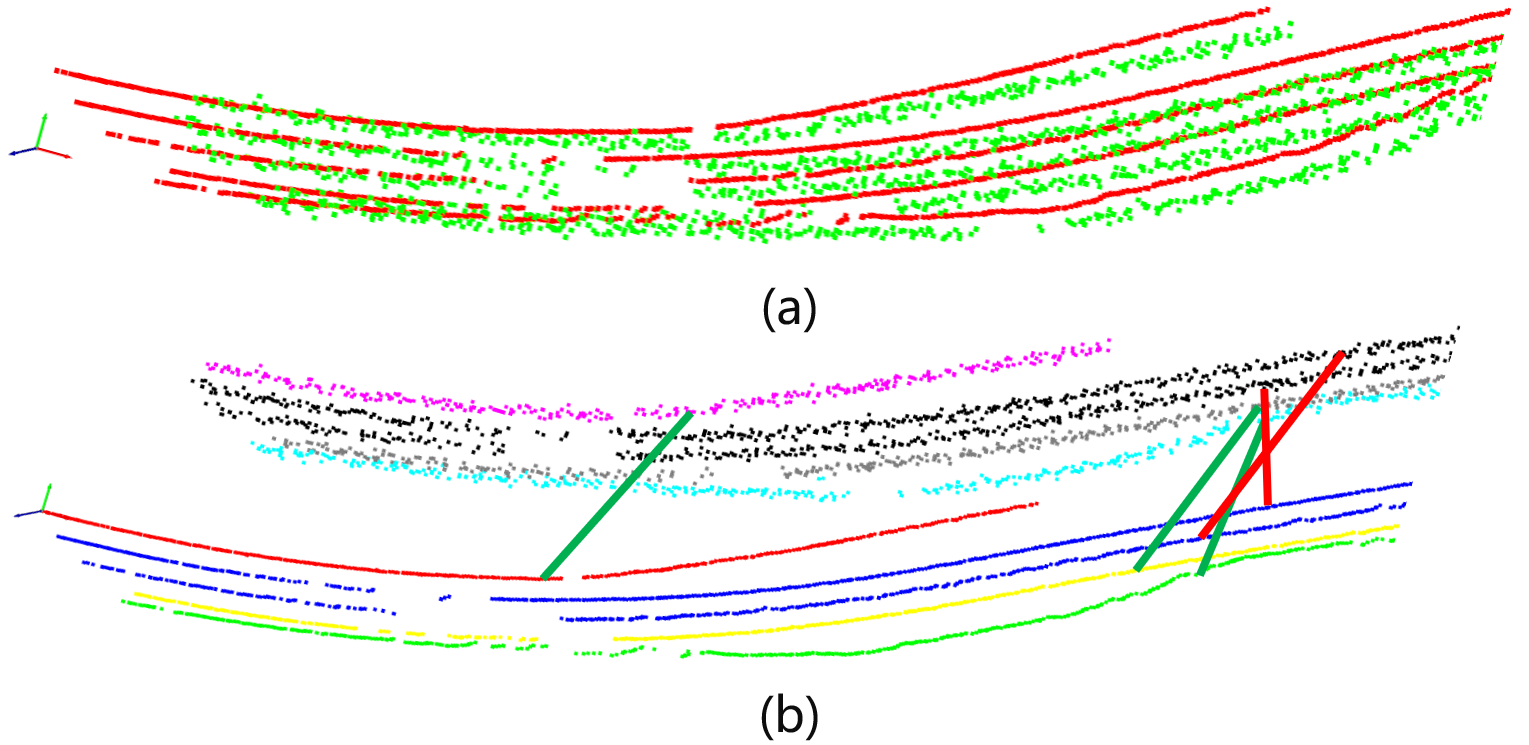}
	\caption{(a) Two frames of lane markings that needed to be associated. (b) Visualization of the association results of these two frames. In each frame, the color represents the lane marking category. In the associations, red represents incorrect, and green represents correct. Due to pose uncertainty, associations based only on Euclidean distances can result in incorrect results.}
	\label{figure:lanes_association}
	\vspace{-0.4cm}
\end{figure}

Nonetheless, relying solely on distance for data association might engender ambiguity, notably in light of the presence of $\Delta R$, as exemplified in Fig. \ref{figure:lanes_association}. Detections and landmarks are not well separated in Euclidean space or even crossed. To this end, we further weighted the edges using lateral order consistency. Just like most graph matching methods~\cite{leordeanu2005spectral}, we define the consistency between two associated pairs (edges in a bipartite graph). To better illustrate the data association, we configure an example in Fig.~\ref{figure:consistency}. There are four lanes, five landmarks, and seven association pairs based on Euclidean distances, three of which are incorrect (but still under the distance upper bound). Each association pair acts as a vertex in the graph. The presence or absence of an edge between two vertices depends on the lateral sequence consistency of the association results they produce. For instance, for $u_1$ and $u_3$ in BEV, we sample two points (e.g., the start and end points) on lane $b_{\prime}$, create a straight line, and determine whether one of the sampled points (e.g., the median point) of lane $a_{\prime}$ is above or below the straight line, again for $a$ and $c$. If they have the same relative relationship, then there is consistency between $u_1$ and $u_3$. In addition, $u_2$ and $u_3$ are also inconsistent since they share landmark $b_{\prime}$. Further, without loss of generality, we take $u_3$ as an example of calculating its consistency score $S(u_3)$.
\begin{equation}
S\left(u_3\right)=\sum_{u_j \in \mathcal{C}\left(u_3\right)} s\left(u_3, u_j\right)
\end{equation}
where $\mathcal{C}$ denotes the set of vertices connected to $u_3$. For $u_3$ and $u_1$, we have,
\begin{equation}
s\left(u_3, u_1\right)=\frac{1}{\operatorname{abs}\left(\phi(a, c)-\phi\left(a^{\prime}, b^{\prime}\right)\right)}
\end{equation}
where $\operatorname{abs}(\cdot)$ denotes the absolute value and $\phi$ denotes the minimum value of the previously mentioned point-to-line distance.

Finally, the edge of the bipartite graph can be derived by multiplying two scores, one being the reciprocal of the previously mentioned Chamfer distance and the other being the corresponding lateral sequence consistency in $S$.

\subsubsection{Pose Update}
\label{subsec:pu}

The problem can be formulated as follows. Let $T_t$ denotes the pose transformation of the camera to the world frame at time $t$. Combined with the association results in Sec.~\ref{subsec:la}, ${ }^d p_k$ is projected to world frame using $T_t$ to determine the parameter $u_k$ on associated spline by finding the foot point $p(u_k)$. Given the local direction $d_k$ of $p(u_k)$, a point-to-tangent residual is defined since the lane marking only provides the lateral constraint for pose estimation. The overall objective of the registration is to find
\begin{equation}
\begin{aligned}
\hat{T}_t=\underset{T_t}{\operatorname{argmin}} \sum_{k=1}^K \rho( \left\|\left(I-d^{\top} d\right)\left(T_t p_k-p\left(u_k\right)\right)\right\|_2^2) &\\
+ \|\left(T_{O_t}\right)^{-1}T_t\|_2^2 &
\end{aligned}
\end{equation}    
where $K$ is the number of points in all detected lanes, $d_k$ can be easily obtained by the first-order derivative of $p(u_k)$, $\rho$ is the robust kernel function, and $T_{O_t}$ is the pose from odometry.

Another critical issue is parameterization. Due to the detection uncertainty and odometry drift, the commonly used parameterization methods, such as the chord length and the centripetal methods, could not work well. To this end, we design a coarse-to-fine parameterization method considering the property of the Catmull-Rom spline: the control points are on the curve, and the segment is restricted between $P_1$ and $P_2$. Specifically, as in Fig.~\ref{figure:parameterization}, we first find the two closest control points and give three conditions:
\begin{enumerate}
    \item These two control points must be sequentially adjacent.
    \item They are neither the first nor the last control point of the spline.
    \item The distance from $p_k$ to the two control points must be less than the distance between the two control points.
\end{enumerate}

Note that since the curvature of the lanes tends to be small for comfort, based on this observation, we assume that the foot point must be on the segment created by the two closest control points. We then sample the three waypoints uniformly on this segment according to $u$ and approximate the curve with a polyline. In this case, the foot point on the curve is converted to the one on the polyline. We determine $u$ according to the foot point.

\subsection{Map Optimization}
\subsubsection{Lane Initialization}
\label{subsec:li}

A good initialization of control points $P$ is necessary before optimizing the splines in the map. To avoid self-intersection, $P$ is expected to be uniformly distributed over the spline with chord length $r$\footnote[1]{We set $r$ to be 3 m.}. To this end, this problem can be described as extending $P$ given detection points $p_{1:M}$.

\begin{figure}[!htbp]
	\centering
	\includegraphics[width=0.85\linewidth]{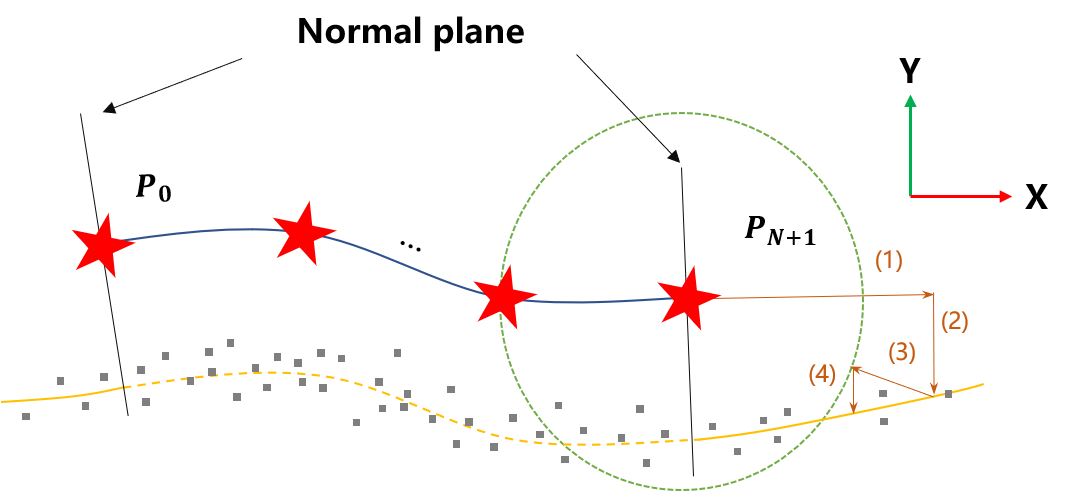}
	\caption{The figure explains how the spline is extended. All data will first be transformed under a local reference frame. Then the normal plane divides the space and selects the observation points for the extension. Finally, the extension is transformed into the problem of finding the intersection of the sphere and the curve along the direction of the spline.}
	\label{figure:extend}
	\vspace{-0.2cm}
\end{figure}

Algorithm 1 describes the main steps of the initialization process. First, if $P$ is empty (new lane marking), we randomly select $p_0$ as a control point. Then, we use $p_{1:M}$ to initialize the candidate set $V$ and keep it updated. To determine $V$, we take the first and last two control points to establish the normal plane, and the normal vector is the tangent of the control point on the spline. As shown in Fig. \ref{figure:extend}, we take only the points beyond these two normal planes and add them to $V$. If not, the algorithm is terminated.

After obtaining $V$, we take $V_0$ to determine whether to extend the head or tail of the spline. Without loss of generality, we take the example of extending to the tail. A simple approach is to draw a sphere with $P$.tail as the center and $r$ as the radius and take the projection of $V_0$ on the sphere as the next control point. This projection has an analytic solution and is easily computed. However, this method is not robust since $p_{1:M}$ can be very noisy and sparse. Instead, we use the LRF and $\mathrm{f}_{xy}$ and $\mathrm{f}_{xz}$ introduced in Sec. \ref{subsec:ldr}. This problem is transformed into the problem of finding the intersection of a sphere and a cubic polynomial curve. However, solving a quintic equation is not easy, so we design a heuristic algorithm as shown in $Expand$ in Algorithm \ref{alg:init}. A more graphic description is shown in Fig. \ref{figure:extend}. Usually, this process ends after three or four iterations. If not, the sphere and the curve have no intersection, and the algorithm updates $P$.tail to $P_s$. It continues to iterate until the condition is met and exits.

\begin{algorithm}[t]
    \label{alg:init}
    \caption{Control Point Initializatin}
    \SetKwInOut{Input}{Input}
    \SetKwInOut{Output}{Output}
    \Input{control points $P$ and detected $p_{1: M}, \mathrm{f}_{xy}, \mathrm{f}_{xz}$}
    \Output{extended $P$}  
    \BlankLine
    \lIf{$P$ is $\emptyset$}{$P$.append($p_0$)}
    $V \leftarrow p_{1:M}$\;
    \While{true}{
        $N_h, N_t \leftarrow ComputeNormalPlane(P_{0:N+1})$\; 
        $V \leftarrow BeyondNormalPlane(V, N_h, N_t)$\;
        \lIf{$V$ is $\emptyset$}{break}
        \eIf{$\|V_0-P$.head$\|_2 \leq \|V_0-P$.head$\|_2$}{
            $Expand(P$.head.next$, P$.head$, \mathrm{f}_{xy}, \mathrm{f}_{xz})$\;
        }{
            $Expand(P$.tail.prev$, P$.tail$, \mathrm{f}_{xy}, \mathrm{f}_{xz})$\;
        }
    }
    \SetKwProg{Def}{def}{:}{}
    \Def{Expand($P_i, P_j, \mathrm{f}_{xy}, \mathrm{f}_{xz}$)}{\label{alg:expand}
        $P_i, P_j \leftarrow LRF.transform(P_i, P_j)$\;
        \eIf{$P_j.x \ge P_i.x$}{
            $P_s=P_j+\left[3r, 0, 0\right]$\;
        }{
            $P_s=P_j+\left[-3r, 0, 0\right]$\;
        }
        \For{$i\leftarrow 1$ \KwTo $10$}{
            $P_f=\left[P_s.x, \mathrm{f}_{xz}(P_s.x), \mathrm{f}_{xz}(P_s.z) \right]$\;
            $P_s=ProjectToSphere(P_f)$\;
            \lIf{$P_s$ changes very little}{break}
        }
        \tcp*[h]{Add $P_s$ to $P$}\;
        $P_s \leftarrow LRF.inv.transform(P_s)$\;
        return $P_s$\;
    }
\end{algorithm}

\subsubsection{Spline Optimization}
\label{subsec:so}
This section will continue to describe how to optimize the control points. We parameterize each detection point $p_k$ based on the association result $(D, L)$ and the optimized pose $\hat{T}_t$ to find its parameter $u_k$. We define the point-to-spline residuals as
\begin{equation}
r_k = \hat{T}_t p_k-p\left(u_k\right)
\end{equation}
where $p\left(u_k\right)$ is a linear function of the control points, and its Jacobian is well found. 

In addition, we need to add another regularization term to the optimization. Although different spline segments share control points, some points still exist that always play the role of $P_0$ and $P_3$. As shown in Fig. \ref{figure:coeffs}, the coefficients of $P_0$ and $P_3$ are always small, which causes the point-to-spline residuals to not constrain these variables well. For this reason, we add constant constraints on the relative relationship between these points and their adjacent points.

The overall factor graph is built as shown in Fig. \ref{figure:system}. To preserve the history information, we use iSAM2~\cite{kaess2012isam2} in GTSAM~\cite{gtsam} for factor graph-based incremental optimization. This allows us to balance speed and accuracy by only updating part of the variables when a new frame comes and not discarding the historical information.

\section{EXPERIMENTS}

Our proposed method is evaluated on OpenLane~\cite{chen2022persformer} lane benchmark, built on the Waymo dataset~\cite{sun2020scalability}. We select this dataset since it has 3D lane marking and instance-level tracking annotations. OpenLane contains 1000 annotated road segments, 798 for training lane detection, and 202 for evaluating the mapping quality and pose estimation. Each segment is approximately 135 meters long on average and has 198 frames at 10 FPS. The entire dataset has 14 annotation categories for lane marks. About 25 percent of the frames contain more than 6 lane markings, which makes lane association very challenging. In addition, the dataset comprises various weather conditions and complex road scenarios such as large curves, up and down slopes, and intersections, which can effectively test the performance of the lane mapping method in the real scenario.

\begin{figure*}[ht]
	\centering
	\subfigure[]{\includegraphics[width=0.43\textwidth]{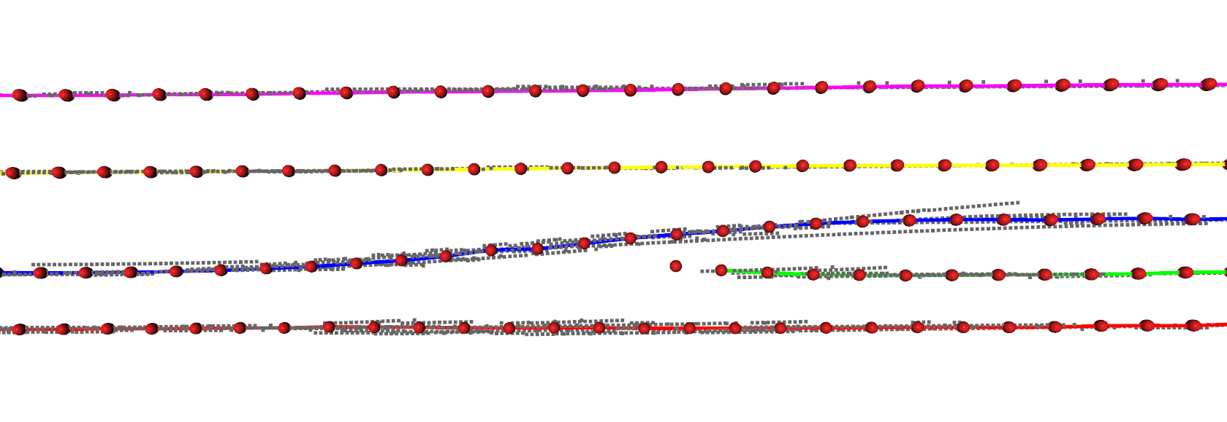}}
    \hspace{0.01\textwidth} 
	\subfigure[]{\includegraphics[width=0.43\textwidth]{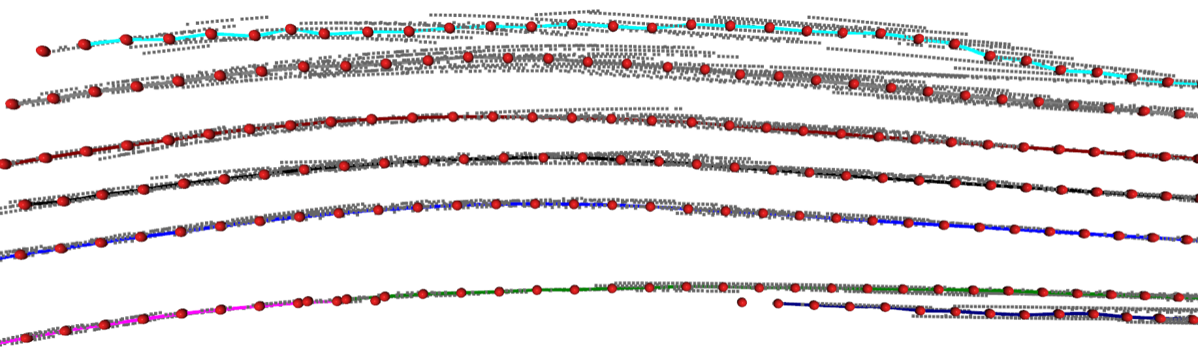}}
    \vspace{-0.01\textwidth}
	\subfigure[]{\includegraphics[width=0.43\textwidth]{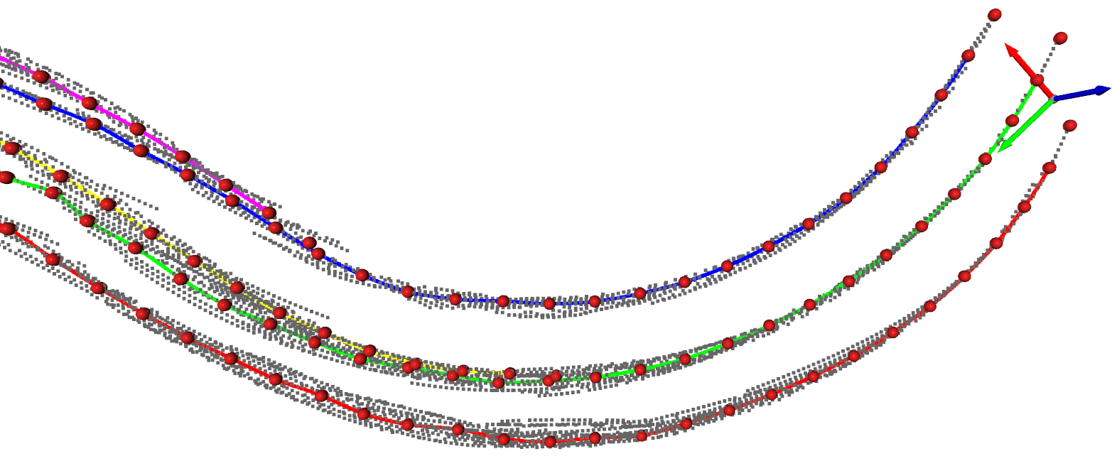}}
    \hspace{0.01\textwidth}
	\subfigure[]{\includegraphics[width=0.43\textwidth]{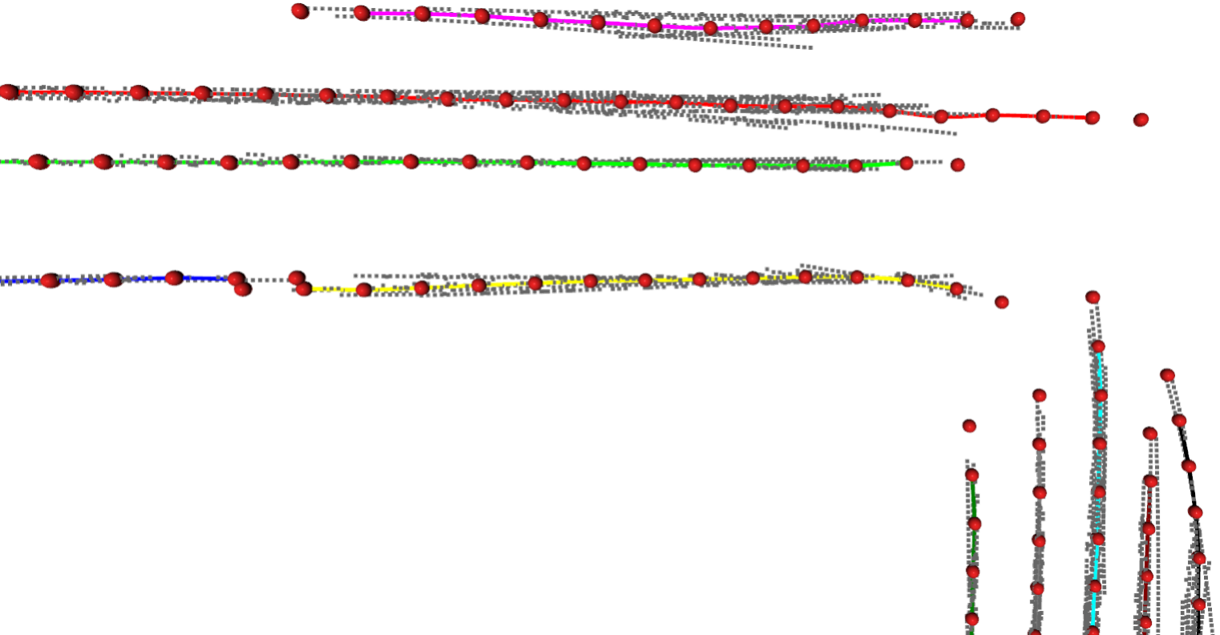}}
    \vspace{-0.01\textwidth}
	\caption{Lane map visualization. Red spheres represent control points, and colored points are sampling points on different spline instances. The gray point cloud is the original detection after downsampling. The figure shows the various scenarios of lanes: split, straight, curved, and intersection.}
    \label{figure:maps}
	\vspace{-0.4cm}
\end{figure*}
\subsection{Evaluation on Lane association}
Thanks to the annotation of the OpenLane dataset for lane tracking, we can obtain the instance IDs of all lane markings in a segment. We take out a pair for association every 10 frames. And to increase the difficulty of data association, we impose a random 3-DOF rigid body pose transformation. The X and Y are subject to the normal distribution $\mathcal{N}(0 \text{ m},3 \text{ m})$, and the yaw is subject to $\mathcal{N}(0^{\circ},2^{\circ})$. The set of pairs with the same lane marking instance ID in both frames is noted as $S_{true}$. For a matching pair given by an association method, it is true positive if it is in $S_{true}$. Otherwise, it is a false positive.

For comparison, we reproduce the shell-based approach proposed in~\cite{kim2021hd} and use the default shell gap 1.5 m~\cite{pannen2019lane}. Also, we do data association based on graph matching and solve it with CLIPPER~\cite{lusk2021clipper}. In CLIPPER, the key issue is to define the point-to-point distance and the noise bound. Our work uses the middle lane marking points to replace itself and finetune the bound. Additionally, we do ablation experiments on our own method. CD+XYZ represents that we omit the lateral sequence consistency. LMR means that the XYZ features are turned into LMR features proposed in~\cite{munoz2022lmr}.
\begin{table}[t]
	\caption{Comparison of Lane Association Performance (\color{red}\bm{$\downarrow$}\color{black}: Lower Better, \color{blue}\bm{$\uparrow$}\color{black}: Higher Better)}
	\vspace{-0.2cm}
	\label{tab:la}
	\centering
	\resizebox{0.485\textwidth}{!}{
        \begin{tabular}{ccccc}
        	\toprule
        	\multicolumn{1}{c}{Method} & \multicolumn{1}{c}{\textbf{F1}\color{blue}\bm{$\uparrow$}} & \multicolumn{1}{c}{\textbf{Precision}$(\%)$\color{blue}\bm{$\uparrow$}} 
            & \multicolumn{1}{c}{\textbf{Recall}$(\%)$\color{blue}\bm{$\uparrow$}} & \multicolumn{1}{c}{\textbf{Time}$(ms)$\color{red}\bm{$\downarrow$}} \\ \midrule
        	Shell\cite{kim2021hd}                      & 0.763 & 80.01 & 74.55 & 15.64 \\
        	CLIPPER\cite{lusk2021clipper}              & 0.795 & \bf{86.77} & 75.66 & 9.027 \\
            CD-XYZ                                     & 0.840 & 84.30 & 83.84 & 1.317 \\
            \makecell{CD-LMR\cite{munoz2022lmr}}       & 0.848 & 85.15 & 84.70 & 27.05 \\ \midrule
        	Ours-LMR\cite{munoz2022lmr}                & 0.946 & 94.89 & 94.62 & \bf{28.85} \\
        	\makecell{Ours}                            & \bf{0.931} & \bf{93.39} & \bf{93.07} & \bf{2.225} \\ \midrule
        	\bottomrule
    	\end{tabular}
    }
	\vspace{-0.4cm}
\end{table}

Tab. \ref{tab:la} demonstrates that our proposed solution is optimal in terms of F1 score and retains a high-efficiency level. It is worth noting that although LMR features are better than using XYZ only, they also introduce a higher computational effort. The original intention of the shell-based approach was to consider Euclidean distances and shape similarities. The significant pose errors make this method ineffective in our experimental setup. In the purely point-based distance methods, CLIPPER is more accurate because of its strict max weighted-clique constraint, but this also reduces the recall.

\subsection{Evaluation on Pose Update}

Since OpenLane does not provide IMU measurement or wheel encoder measurement, this prevents us from running the VIO algorithm to obtain the actual odometry pose. To simulate the drift caused by the odometry, we add a random noise of 3-DOF between every two frames. The added noise is much larger than most odometry works performed on the KITTI odometry benchmark~\cite{geiger2012we}.

\begin{table}[h]
	\caption{Evaluation of pose update performance (\textcolor{blue}{updated}/\textcolor{red}{odometry})}
	\label{tab:pu}
	\centering
	\resizebox{0.485\textwidth}{!}{
        \begin{tabular}{ccccccc}
        	\toprule
        	\multicolumn{1}{c}{} & \multicolumn{2}{c}{\textbf{10m}} & \multicolumn{2}{c}{\textbf{30m}} & \multicolumn{2}{c}{\textbf{50m}} \\ 
        	\multirow{-2}{*}{Noise} & $Rot (^\circ)$\color{red}\bm{$\downarrow$} & $Trans (m)$\color{red}\bm{$\downarrow$} & $Rot (^\circ)$\color{red}\bm{$\downarrow$} & $Trans (m)$\color{red}\bm{$\downarrow$} & $Rot (^\circ)$\color{red}\bm{$\downarrow$} & $Trans (m)$\color{red}\bm{$\downarrow$} \\ \midrule
        	$(0.1,0.1)$   & 0.415/0.460  & 0.498/0.534  & 0.620/0.654  & 0.794/0.802  & 0.753/0.804  & 0.997/1.022 \\
        	$(0.2,0.2)$   & 0.687/0.921  & 0.866/1.068  & 0.994/1.308  & 1.358/1.604  & 1.215/1.609  & 1.657/2.044 \\
        	$(0.3,0.3)$   & 0.944/1.381  & 1.224/1.603  & 1.331/1.962  & 1.893/2.407  & 1.568/2.413  & 2.278/3.067 \\
        	$(0.4,0.4)$   & 1.178/1.841  & 1.632/2.137  & 1.601/2.616  & 2.460/3.210  & 1.884/3.217  & 2.928/4.091 \\
        	$(0.5,0.5)$   & 1.464/2.302  & 2.088/2.672  & 1.980/3.271  & 3.127/4.013  & 2.291/4.022  & 3.677/5.116 \\ \midrule
        	\bottomrule
    	\end{tabular}
    }
	\vspace{-0.2cm}
\end{table}

Tab. \ref{tab:pu} demonstrates the role of lane markings in pose estimation. The first column illustrates the added noise level, e.g., (0.1, 0.1) represents the rotational noise standard deviation of 0.1° and the translational noise standard deviation of 0.1 m. We calculate the relative pose error~\cite{geiger2012we} in the distance 10 m, 30 m, and 50 m, respectively. The right side of the slash is the result of the original odometry with noise added. In contrast, the left side represents the updated pose. The results in the table show that the overall lane markings can help improve odometry-only pose estimation. However, the improvement is little when the noise is slight. This reveals a possible upper bound for the lane marking's improvement in the pose estimation. It might be due to two reasons: the detected 3D lane markings from networks are not so accurate, and the uncertainty of splines is not modeled very well in the map.

\subsection{Evaluation on Map Quality}

OpenLane does not provide the ground truth of the global HD map but only 3D lane marking annotations for each frame. Thus we project the spline from the map onto each frame and perform point sampling. For a fair comparison, we do not project the global map to each frame but only the online local map constructed with historical information. Like the evaluation in OpenLane~\cite{chen2022persformer}, we report the recall and precision for each frame within a local area of 50 m\footnote[1]{In OpenLane, it's 100 m. We choose 50 m because we only use lane detection within 50 m for mapping.} ahead. A point on a lane marking becomes valid if the distance between it and a point on the ground truth is less than 0.5 m\footnote[2]{This value defaults to 1.5 m in OpenLane. In addition, we use KDTree to find the nearest points instead of sampling along a fixed Y-axis interval because we see some Y-intervals without lane marking points.}. True positive is defined as a lane marking where the number of valid points exceeds 75$\%$ of the corresponding GT. 

\begin{table}[h]
	\caption{Comparison of Lane Mapping F1 Score \color{blue}\bm{$\uparrow$}}
	\vspace{-0.2cm}
	\label{tab:lmf}
	\centering
	\resizebox{0.485\textwidth}{!}{
        \begin{tabular}{cccccccc}
        	\toprule
        	\multicolumn{1}{c}{Method} & \multicolumn{1}{c}{\textbf{Up $\&$ Down}} & \multicolumn{1}{c}{\textbf{Curve}} & \multicolumn{1}{c}{\textbf{Extreme Weather}} & \multicolumn{1}{c}{\textbf{Intersection}} & \multicolumn{1}{c}{\textbf{Merge $\&$ Split}} & \multicolumn{1}{c}{\textbf{Night}} & \multicolumn{1}{c}{\textbf{All}} \\ \midrule
        	PersFormer\cite{chen2022persformer}    & 0.407 & 0.595 & 0.514 & 0.474 & 0.575 & 0.518 & 0.559 \\
        	Ours                                   & \bf{0.459} & \bf{0.609} & \bf{0.552} & \bf{0.523} & \bf{0.631} & \bf{0.566} & \bf{0.609} \\  \midrule
        	\bottomrule
    	\end{tabular}
    }
	\vspace{-0.2cm}
\end{table}

\begin{table}[h]
	\caption{Comparison of Lane Mapping on All Validation Segments}
	\vspace{-0.2cm}
	\label{tab:lmavs}
	\centering
	\resizebox{0.45\textwidth}{!}{
        \begin{tabular}{ccccc}
        	\toprule
        	\multicolumn{1}{c}{Method} & \multicolumn{1}{c}{\textbf{F1}\color{blue}\bm{$\uparrow$}} & \multicolumn{1}{c}{\textbf{Recall}\color{blue}\bm{$\uparrow$}} 
            & \multicolumn{1}{c}{\textbf{Precision}\color{blue}\bm{$\uparrow$}} & \multicolumn{1}{c}{\textbf{F1 increase}\color{blue}\bm{$\uparrow$}} \\ \midrule
        	PersFormer (0.8)                   & 0.497 & 0.418 & 0.614 & - \\
        	PersFormer (0.6)                   & 0.515 & 0.445 & 0.613 & - \\
        	PersFormer (0.4)                   & 0.530 & 0.467 & 0.611 & - \\
        	PersFormer (0.0)                   & 0.559 & 0.517 & 0.609 & - \\ \midrule
        	Ours (0.8)                         & 0.539 & 0.450 & 0.672 & $8.45\%$ \\
        	Ours (0.6)                         & 0.559 & 0.479 & 0.670 & $8.54\%$ \\
        	Ours (0.4)                         & 0.576 & 0.505 & 0.670 & $8.68\%$ \\
        	Ours (0.0)                         & \bf{0.609} & \bf{0.559} & \bf{0.668} & $8.94\%$ \\  \midrule
        	\bottomrule
    	\end{tabular}
    }
	\vspace{-0.5cm}
\end{table}
To evaluate, more precisely, the performance of the proposed method, we divided the validation set into different scenarios. The results in Tab. \ref{tab:lmf} show that our proposed method improves the recall and precision of lane markings by leveraging past detection in all scenarios. However, we note that the improvement is minimal in the curve scenario. By comparing the original detection with the optimized splines, we found that this scenario results in many errors for large curvature, especially at a distance. This may be related to the fact that the detector we use is based on straight anchors. In this case, the large number of outliers creates disturbances for the optimizer.

Also, to evaluate the degree of improvement of our system for different performance detectors, we randomly remove a lane marking from the current detection with $p$ probability. As shown in Tab. \ref{tab:lmavs}, the maximum $p$ is 0.8. The results demonstrate that our proposed method can steadily improve the quality of lane markings in the map under different levels of interference.

\section{Discussion and Conclusion}
In contrast to single-frame online mapping, the proposed system is capable of leveraging historical detection. Retaining historical information facilitates the enhancement of lane marking recall. Furthermore, incorporating historical observations into the optimization yields a more precise representation of lane markings. Additionally, the system directly generates a vectorized map embodied by splines. Nevertheless, the method's performance remains contingent upon lane detection performance. Moreover, modeling the uncertainty of network predictions necessitates effort, as opposed to physical sensors. Lastly, loop closure proves essential for constructing a globally consistent lane map. In subsequent research, we shall persist in exploring online lane mapping to address the aforementioned challenges.

\bibliographystyle{IEEEtran}
\bibliography{root}
\end{document}